\documentclass[lettersize,journal]{IEEEtran}
\usepackage{amsmath,amsfonts}
\usepackage{newtxtext,newtxmath}
\usepackage{times} 
\usepackage{algpseudocode}
\usepackage{algorithm}
\usepackage{array}
\usepackage{textcomp}
\usepackage{stfloats}
\usepackage{url}
\usepackage{verbatim}
\usepackage{graphicx}
\usepackage{cite}
\usepackage{subcaption}  
\usepackage{listings}
\usepackage{url}
\usepackage{booktabs} 
\usepackage{threeparttable} 
\usepackage{lscape} 
\usepackage{caption} 
\usepackage{booktabs}
\usepackage{booktabs}
\usepackage{multirow}
\usepackage{tabularx}
\usepackage{makecell}

\begin{document}

\title{Explosive Output to Enhance Jumping Ability: A Variable Reduction Ratio Design Paradigm for Humanoid Robots Knee Joint}

\author{Xiaoshuai Ma, Haoxiang Qi, Qingqing Li, Haochen Xu, Xuechao Chen, Junyao Gao, Zhangguo Yu, Qiang Huang 
\thanks{This paper was produced by the IEEE Publication Technology Group. They are in Piscataway, NJ.}
\thanks{Manuscript received April 19, 2021; revised August 16, 2021.}}

\markboth{IEEE ROBOTICS AND AUTOMATION LETTERS,~Vol.~14, No.~8, August~2025}%
{Shell \MakeLowercase{\textit{et al.}}: A Sample Article Using IEEEtran.cls for IEEE Journals}


\maketitle

\begin{abstract}
Enhancing the explosive power output of the knee joints is critical for improving the agility and obstacle-crossing capabilities of humanoid robots. However, a mismatch between the knee-to-center-of-mass (CoM) transmission ratio and jumping demands, coupled with motor performance degradation at high speeds, restricts the duration of high-power output and limits jump performance. To address these problems, this paper introduces a novel knee joint design paradigm employing a dynamically decreasing reduction ratio for explosive output during jump. Analysis of motor output characteristics and knee kinematics during jumping inspired a coupling strategy in which the reduction ratio gradually decreases as the joint extends. A high initial ratio rapidly increases torque at jump initiation, while its gradual reduction minimizes motor speed increments and power losses, thereby maintaining sustained high-power output. A compact and efficient linear actuator-driven guide-rod mechanism realizes this coupling strategy, supported by parameter optimization guided by explosive jump control strategies. Experimental validation demonstrated a 63 cm vertical jump on a single-joint platform (a theoretical improvement of 28.1\% over the optimal fixed-ratio joints). Integrated into a humanoid robot, the proposed design enabled a 1.1 m long jump, a 0.5 m vertical jump, and a 0.5 m box jump.

\end{abstract}

\begin{IEEEkeywords}
   Humanoid robots, Explosive jumping, Variable reduction, Knee joint design, Linear actuator
\end{IEEEkeywords}

\section{Introduction}

\IEEEPARstart{J}{umping} is a fundamental yet challenging form of locomotion for humanoid robots, playing a crucial role in enabling rapid traversal of complex terrains, overcoming large obstacles, and enhancing overall mobility \cite{zhang2020biologically,dobra2020technology,armour2007jumping}. 
Recent advances in actuation and control have greatly improved robotic performance \cite{PengjieXiang1,Pengjie2}. However, humanoid robots still fall far short of human-level jumping ability. Currently, the Boston Dynamics Atlas robot, which is hydraulically actuated, demonstrates the most advanced jumping capabilities, achieving vertical jumps of over 0.6 meters and performing complex dynamic maneuvers such as agile parkour \cite{bostondynamics2023atlas}. However, hydraulic systems face challenges including leakage, operational noise, high cost, and excessive weight, and Boston Dynamics has since discontinued research on the hydraulic version \cite{toombes2011numerical}.

As the current mainstream approach, electrically driven humanoid robots, which are primarily actuated by permanent magnet synchronous motors (PMSMs), offer advantages such as high efficiency, precise control, low noise, and lightweight design \cite{Yuchuang}. However, as shown in Table~\ref{tab:robot_comparison}, their jumping capabilities remain considerably lower than those of hydraulic robots and humans.

During jumping, the robot must rapidly accumulate sufficient kinetic energy within 0.15 to 0.5 seconds \cite{Domire}, which places extremely high demands on the energy output of the leg joints, particularly the knee joint \cite{qi2023vertical,chen2021motion,jiang2018motion,toshihide2024joint}. However, high-power motors are typically large and heavy, increasing the size and rotational inertia of the knee joint and consequently reducing the flexibility and responsiveness of the leg. Therefore, developing a knee joint that is both compact and lightweight yet capable of delivering explosive power remains a significant challenge.

Electrically driven knee joints fall into two main categories: rotary actuators and linear actuators. Rotary actuators combine a motor with a planetary gearbox, a harmonic drive, or a cycloidal drive. Although planetary gearboxes are compact, efficient, and capable of high loads at reasonable cost~\cite{wensing2017proprioceptive,Sciarra}, their relatively large mass increases leg inertia. To boost knee torque without adding inertia, solutions such as Unitree H1~\cite{Unitree2024ow} and electric Atlas~\cite{bostondynamics_electric} employ high-torque motors mounted near the hip and transmit power through linkages. Unitree H1 achieves 350\,Nm peak torque and can perform backflips~\cite{unitree2024_Backflip}. Other platforms, including Unitree G1~\cite{G1jump03}, Engine AI~\cite{EngineAI2025}, and the MIT humanoid~\cite{chignoli2021humanoid}, reduce overall robot size and mass to lower knee loading and improve agility, enabling martial arts \cite{Unitree_G1_2025_kufu}, flips \cite{unitree2024_Backflip}, and a 0.5 m vertical jump \cite{chignoli2021humanoid} with significant leg retraction. The cable-driven JAXON-3P~\cite{kojima2020drive} uses a harmonic drive and tendon transmission to halve leg mass and reach 700\,Nm peak torque but achieves only a 0.3\,m vertical jump. Although these designs enhance dynamic performance, their explosive jumping capability remains limited.

The transmission ratio from the knee joint to the center of mas (CoM) increases with joint extension (Fig. 1), which conflicts with the jumping requirement for high torque followed by high speed \cite{armour2007jumping}. Furthermore, motor torque is limited at low speeds and power output is insufficient at high speeds, exacerbating this mismatch. Since rotary actuator joints have fixed or only slightly varying reduction ratios, they cannot address this issue. As a result, the knee motor’s performance cannot be fully exploited during jumping, leading to poor jumping capability.

In addition, Cassie adopts parallel RV reducers and integrates elastic elements to reduce motor speed demands, enabling vertical jumps of 0.5 meters, forward jumps of 1.4 meters, and box jumps of 0.4 meters \cite{li2023robust}. Nevertheless, its large leg volume limits terrain adaptability, and adding an upper body significantly increases weight, degrading overall jumping performance \cite{AgilityRobotics}.

Linear actuators typically integrate PMSMs with roller screws or ball screws. Robots such as Optimus \cite{TeslaAI}, Lola \cite{lohmeier2010design}, and RH5 \cite{ebetaer2021design} utilize linear actuators and demonstrate stable walking. However, their designs primarily focus on torque output rather than dynamic motion, limiting their capability for explosive jumping.

Moreover, small-scale robots enhance jump performance through integrated parallel springs \cite{hong2020combined}, parallel joint mechanisms \cite{gao2023alternating}, and momentum-boosting techniques \cite{zhao2015msu}. While effective in lightweight systems, these methods face significant challenges when scaled to full-sized humanoids, due to increased structural complexity, extended transmission chains, limited energy density of elastic components, and substantial control difficulties. 

This paper presents the design paradigm of an electrically driven humanoid knee joint to enhance jumping ability. By analyzing the knee’s kinematic characteristics and the motor’s output capability during the takeoff phase, we establish a coupling between joint angle and reduction ratio to boost motor output during takeoff and thus improve jumping performance. Based on this concept, we propose a practical structural implementation and a parameter optimization method guided by jump control strategies. The effectiveness of the approach is validated through numerical simulations, experiments on a single-leg test platform, and trials with a full-scale humanoid robot.

The major contributions of this paper are as follows.
\begin{itemize}
    \item A novel explosive variable reduction ratio knee (EVRR-K) strategy  for enhancing jumping ability, in which the reduction ratio decreases progressively with joint extension, enabling sustained high-power output from the knee joint motor throughout the jumping phase. 
    \item A compact and efficient knee mechanism is introduced to implement the proposed EVRR-K strategy. Driven by a linear actuator, the mechanism features a simple structure and high parametric flexibility, enabling the realization of diverse reduction ratio curves.
    \item A novel parameter optimization method guided by explosive jump control strategies is introduced to identify the optimal reduction ratio curve for improved jumping performance.
\end{itemize}

The remainder of this paper is organized as follows. Section~\ref{sec2} presents the EVRR-K strategy. Section~\ref{sec3} proposes the corresponding structural implementation. Section~\ref{sec4} introduces a parameter optimization method for EVRR-K. Section~\ref{sec5} describes the experimental validation. Finally, Section~\ref{sec6} concludes the paper and outlines future research directions.

\begin{table}[t]
    \centering
    \caption{Specifications and Jumping Performance of  electrically driven Humanoid Robots}
    \label{tab:robot_comparison}
    \resizebox{\columnwidth}{!}{
    \begin{threeparttable}
    \begin{tabular}{lccc}
    \toprule
    \textbf{Robot} & \textbf{Mass/Height} & \textbf{Knee Peak Torque} & \textbf{Jump Performance} \\
     & (kg/m) & (Nm) & Height/forward /Box (m) \\
    \midrule
    Unitree H1 \cite{Unitree2024ow} & 47/1.8 & 360 & 0.3\tnote{*} /--/-- \\ 
    Unitree G1 \cite{G1jump03} & 35/1.3 & 120 & 0.3\tnote{*} /1.4/-- \\ 
    MIT-Humanoid \cite{chignoli2021humanoid} & 24/1.04 & 144 & 0.5/--/-- \\ 
    JAXON 3-P \cite{kojima2020drive} & 35/1.7 & 700 & 0.3/--/-- \\ 
    Cassie \cite{li2023robust} & 27.2/1.3 & -- & 0.5/1.4/0.4 \\
    BHR8-J1 (ours) & 45/1.6 & 318 & 0.5/1.1/0.5 \\  
    \bottomrule
    \end{tabular}
    \begin{tablenotes}
    \small
    \item [*] Estimated values based on published specifications.
    \item [--] Indicates no publicly available data.
    \end{tablenotes}
    \end{threeparttable}
    }
\end{table}

\section{EVRR-K STRATEGY FOR EXPLOSIVE JUMPS}\label{sec2} 
To improve knee output during takeoff, we analyze both motor output characteristics and knee kinematics, and propose a heuristic strategy that couples the reduction ratio to the joint angle.

\subsection{Motor Output Characteristics Analysis}
The PMSM is the primary actuator in legged robots, and its peak output capability is mainly limited by power losses and thermal constraints. The total power loss of the motor comprises copper loss, iron loss, and mechanical loss, as given by:
\begin{equation}
P_{\text{loss}} = P_{\text{copper}} + P_{\text{iron}} + P_{\text{mechanical}}
\end{equation}
where $P_{\text{copper}}$ denotes the winding resistance loss (related to current), $P_{\text{iron}}$ represents hysteresis and eddy current losses (related to speed), and $P_{\text{mechanical}}$ refers to friction and windage losses (also related to speed).

\begin{figure}[h]
    \centering
    \includegraphics{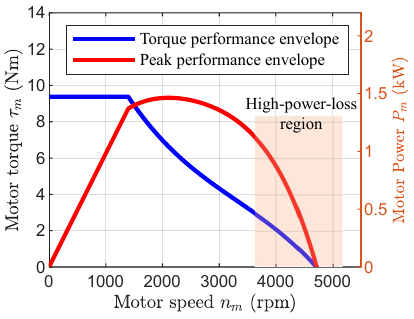}
    \caption{Motor torque performance envelope (TPE) and power performance envelope (PPE), with the high-power-loss region highlighted.}
    \label{fig:high_power_loss_region}
\end{figure}

Fig.~\ref{fig:high_power_loss_region} shows the motor performance envelope. At low speeds, peak output is thermally limited by the maximum allowable current, yielding a constant peak torque. In the mid-speed region, output is constrained by the rated power, resulting in an approximately constant peak power. At high speeds, power losses rise rapidly, causing a steep decline in output power. Therefore, to maximize the motor’s output during jumping, it is best to avoid thermal and excessive power-loss constraints by operating near the mid-speed region.

Additionally, the relationship between joint output power and motor output power is given by:
\begin{equation}
    P_{\text{J}} = \eta_{\text{J}} P_{\text{m}}
\end{equation}
where \( P_{\text{J}} \) is the joint output power, \( P_{\text{m}} \) is the motor power, and \( \eta_{\text{J}} \) represents the transmission efficiency, which is influenced by the reduction mechanism.

\subsection{Kinematic Analysis of the Knee Joint During Jumping Based on a Simplified Model}\label{sec2.2}

\begin{figure}[!h]
    \centering
    \includegraphics{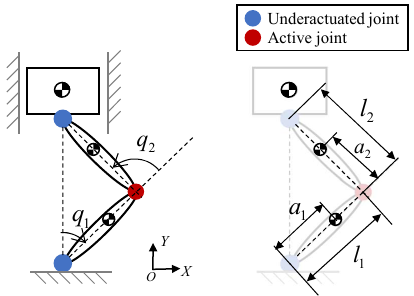}
    \caption{Prototype system used to analyze knee joint requirements in jumping.}
    \label{fig:leg_model}
\end{figure}

To systematically analyze the dynamic requirements of the knee joint during jumping, a simplified leg model is employed to capture the primary motion characteristics. As illustrated in Fig.~\ref{fig:leg_model}, this model considers the knee as the main actuated joint, while the hip and ankle are treated as passive and constrained to vertical motion. This simplification enables focused analysis of the knee joint's contribution to the propulsion of the center of mass (CoM) during takeoff.

Takeoff is along the $y$-axis; the CoM velocity in this direction is:
\begin{equation}\label{eq:ycom}
\dot y_{\mathrm{CoM}}
= J_{\mathrm{CoM},y}(q_2)\,\dot q_2,
\quad
J_{\mathrm{CoM},y}(q_2)
= \frac{\partial\,y_{\mathrm{CoM}}(q_2)}{\partial q_2}\,
\end{equation}
where $\dot y_{\mathrm{CoM}}$ represents the velocity of the CoM in the $y$ direction, where $J_{\mathrm{CoM},y}(q_2)$ is the Jacobian component of the CoM with respect to the knee joint angle $q_2$, and $\dot q_2$ is the angular velocity of the knee joint.

Therefore, the transmission ratio from the knee joint to the center of mass along the y‐axis can be expressed as
\begin{equation}\label{eq:effective_ratio}
\lambda(q_2)
= \frac{\dot q_2}{\dot y_{\mathrm{CoM}}}
= \frac{1}{J_{\mathrm{CoM},y}(q_2)}\,
\end{equation}
where the Jacobian $J_{CoM,y}(q_2)$ is given by:
\begin{equation}\label{eq:Jcy2}
{J_{CoM,y}(q_2)} = \frac{[a_1 m_1 + (l_1 + a_2)m_2 + (l_1 + l_2)m_3]\sin(q_2/2)}{m_1 + m_2 + m_3}
\end{equation}  
where, $a_1$ and $a_2$ represent the distances from the centers of mass of the shank and thigh to the previous joint, $l_1$ and $l_2$ are the lengths of the shank and thigh, and $m_1$, $m_2$, and $m_3$ denote the masses of the shank, thigh, and torso, respectively.

\begin{figure}[!h]
    \centering
    \begin{subfigure}{0.23\textwidth}
        \centering
        \includegraphics{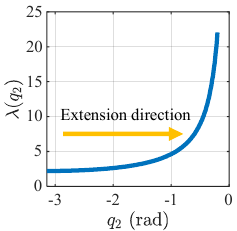}
        \caption{}\label{fig:general_ratio}
    \end{subfigure}
    \hfill
    \begin{subfigure}{0.23\textwidth}
        \centering
        \includegraphics{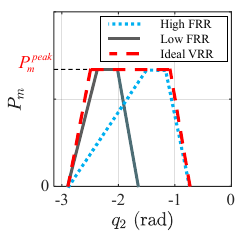}
        \caption{}\label{fig:idealVRR}
    \end{subfigure}
    \caption{(a)Knee-CoM y-axis transmission ratio $\lambda(q_2)$ as a function of knee joint angle $q_2$; (b) Ideal transmission ratio curve for motor explosive output.}
    \label{fig:ideal_ratio_and_general_ratio}
\end{figure}

According to Eqs.~\ref{eq:effective_ratio} and \ref{eq:Jcy2}, the transmission ratio $\lambda(q_2)$ increases with the knee joint angle $q_2$ (Fig.~\ref{fig:general_ratio}), and its rate of increase also grows with $q_2$. Once $q_2$ exceeds $-1\,$rad, $\lambda(q_2)$ rises sharply.

The CoM force and velocity along the $y$-axis are
\begin{equation}\label{eq:Fcom}
F_{\mathrm{CoM},y} = \tau_{2}\,\lambda(q_2),
\quad
\dot y_{\mathrm{CoM}} = \dot q_{2}\,\lambda(q_2),
\end{equation}
where $\tau_{2}$ is the knee joint torque.

The fixed reduction ratio knee (FRR-K) exhibits a trade-off: a low ratio causes insufficient takeoff acceleration and delayed entry into the high-power region, while a high ratio requires impractical motor speeds late in the jump and fails to sustain high-power output (Fig.~\ref{fig:idealVRR}). To overcome this, we propose a EVRR-K strategy with a ratio progressively decreasing during extension. This enables rapid entry into and maintenance within the high-power region at takeoff, significantly boosting jump performance (Fig.~\ref{fig:idealVRR}).

Furthermore, as Eq.~\ref{eq:Jcy2}, increasing the CoM height and limb lengths raises $J_{com,y}(q_2)$, which lowers $\lambda(q_2)$ and the required motor speed, thereby reducing leg inertia and improving dynamic performance~\cite{wensing2017proprioceptive}. These parameters should be chosen to match the overall robot design requirements.

\section{MECHANICAL DESIGN}\label{sec3}
\begin{figure}[!h]
    \centering
    \includegraphics{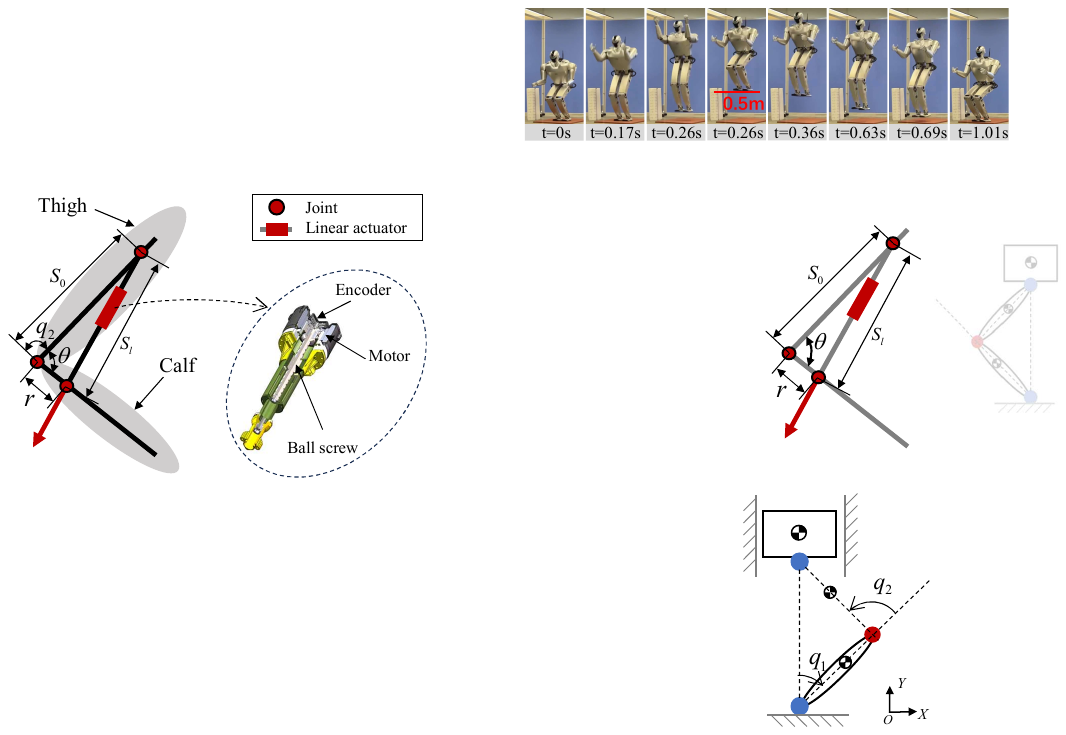}
    \caption{Schematic of the high-explosiveness variable reduction ratio knee joint and high-efficiency linear actuator.}
    \label{fig:high_explosive_joint}
\end{figure}

To realize the above strategy, a linear actuator-driven guide-rod mechanism is employed to generate knee joint rotation, as illustrated in Fig.~\ref{fig:high_explosive_joint}. To further enhance joint output capability, a high-efficiency, dual-lead ball screw with a 10~mm lead is selected~\cite{rigacci2020experimental}. The relationship between the reduction ratio and the joint angle is given by:
\begin{equation}\label{eq:k_joint}
    k  = \frac{2\pi r (S_0 + r) \sin \theta}{Q \sqrt{2S_0 r - 2 r^2 \cos \theta + S_0^2 + 2r^2 - 2 S_0 r \cos \theta}}
\end{equation}
where $Q$ is the ball screw lead, $r$ is the crank length, $\theta$ is the angle between the crank and the frame, and $S_0$ is the frame length.
The relationship between the knee joint angle $q_2$ and $\theta$ can be expressed as:
\begin{equation}
    q_2 = \theta - \pi
\end{equation}

\begin{figure}[!h]
    \centering
       \begin{subfigure}{0.23\textwidth}
        \centering
        \includegraphics[width=\textwidth]{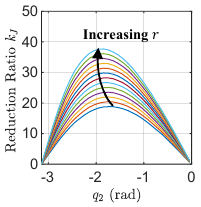}
        \caption{Effect of $r$. }\label{fig:joint_diagram}
    \end{subfigure}
    \hfill
    \begin{subfigure}{0.23\textwidth}
        \centering
        \includegraphics[width=\textwidth]{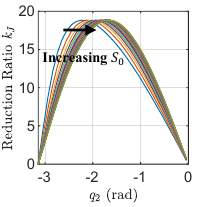}
        \caption{Effect of $S_0$.}\label{fig:lactuator_structure}
    \end{subfigure}
    \caption{Effect of variations in structural parameters on the coupling between joint angle and reduction ratio ($Q = 10$~mm).  (a) $S_0 = 250$~mm, $r$ varies from 10 to 50~mm; (b) $r = 30$~mm, $S_0$ varies from 150 to 250~mm.}
    \label{fig:high_explosive_joint1}
\end{figure}

According to Eq.~\ref{eq:k_joint}, Fig.~\ref{fig:high_explosive_joint1} illustrates how design parameters affect the coupling between reduction ratio and joint angle. Regardless of parameter values, the reduction ratio first increases and then decreases as the joint extends. Varying the link length $r$ changes the peak reduction ratio (Fig.~\ref{fig:joint_diagram}), while the minimum remains zero. Adjusting the offset $S_0$ shifts the joint angle at which the maximum reduction ratio occurs within the range $[-\pi, -\pi/2]$ (Fig.~\ref{fig:lactuator_structure}). These results demonstrate that the mechanism offers wide parameter tunability, enabling a decreasing reduction ratio in the extended pose and fully meeting the kinematic requirements of this study.

Additionally, the coupling between joint angle and reduction ratio can be further tuned by adjusting the angle between the hinge connection line and the joint axis, thereby increasing design flexibility. In this case, $q_2$ can be expressed as:
\begin{equation}
    q_2 = \theta - \pi + \Delta \theta
\end{equation}
where $\Delta \theta$ denotes the angular offset introduced during initial assembly.

\section{JOINT PARAMETER OPTIMIZATION BASED ON EXPLOSIVE JUMPING CONTROL}\label{sec4}
In this section, we optimize the knee joint structure by developing a parameter optimization framework based on the simplified model. The proposed approach is also compared with an optimal fixed-ratio scheme.

\subsection{Optimization Method}
\begin{figure} 
    \centering
    \includegraphics{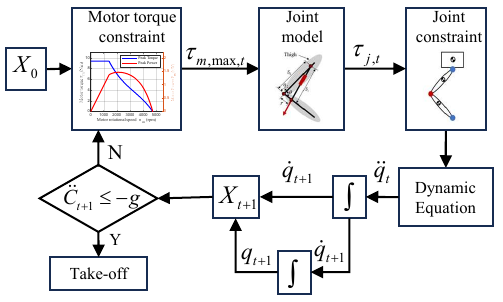}
    \caption{Explosive Jump Control Process}
    \label{fig:optimization_method}
  \end{figure}
  
To maximize the knee joint motor output during takeoff, we implement a explosive control strategy~\cite{jiang2018motion}, as illustrated in Fig.~\ref{fig:optimization_method}, which drives the joint to operate at its maximum available torque. The maximum torque that the motor can provide is given by:
\begin{equation}
    {\tau _{m,max}} = \left\{ {\begin{array}{*{20}{c}}
    {{\tau _{\text{peak}}},} & {\omega_m \leq \omega_{\text{break}}} \\ 
    {{\tau _{\text{limit}}(I_q, \omega_m),}} & {\omega_m \in (\omega_{\text{break}}, \omega_{\text{max}}]}
    \end{array}} \right.
\end{equation}
where \( \tau_{m} \) represents the motor torque output during control, \( \tau_{\text{peak}} \) is the peak torque of the motor, and \( \omega_m \) is the motor angular velocity, \(I_q\) is the motor’s \(q\)-axis current. \( \omega_{\text{break}} \) denotes the speed at which the motor enters the power-limited region, where torque starts decreasing with speed. \( \omega_{\text{max}} \) is the theoretical maximum motor speed. Additionally, \( \tau_{\text{limit}}(I_q, \omega_m) \) represents the maximum torque that can be delivered in the power-limited region.

The joint torque output can be expressed as:

\begin{equation}
    {\tau _{J}} = {\tau _{m }}{k_J}{(q_{2})}
\end{equation}
where \( k_J(q_2) \) is the joint transmission ratio at a given knee joint angle \( q_2 \).

The velocity of the robot’s CoM is given by:

\begin{equation}
    {{\dot y}_{CoM,t}} = J({q_{2,t}}){{\dot q}_{2,t}}
\end{equation}

When the CoM acceleration falls below \(-g\), the robot enters free flight. The CoM acceleration is given by
\begin{equation}\label{eq:com_accel}
\ddot y_{\mathrm{CoM}}(t)
= \dot J\bigl(q_2(t)\bigr)\,\dot q_2(t)
+ J\bigl(q_2(t)\bigr)\,\ddot q_2(t)\,,
\end{equation}
where \(\dot J\) denotes the time derivative of the Jacobian.

At takeoff (\(t = t_{\mathrm{to}}\)), the total mechanical energy is the sum of kinetic and potential components:
\begin{equation}\label{eq:W_takeoff}
W_{\mathrm{takeoff}}
= \frac{1}{2}\,m_{\mathrm{tot}}\,\dot y_{\mathrm{CoM}}(t_{\mathrm{to}})^2
+ m_{\mathrm{tot}}\,g\,y_{\mathrm{CoM}}(t_{\mathrm{to}})\,.
\end{equation}

To maximize the robot’s jump height, we maximize this takeoff energy:
\begin{equation}\label{eq:opt_max}
\max_{r, S_0, \Delta\theta}\; W_{\mathrm{takeoff}}\,,
\end{equation}
which is equivalent to
\begin{equation}\label{eq:opt_min}
\min_{r, S_0, \Delta\theta}\; f(r, S_0, \Delta\theta)
= -\,W_{\mathrm{takeoff}}\,,
\end{equation}
where \(r\), \(S_0\) and \(\Delta\theta\) are the design parameters of the VRR joint.






The design parameters are subject to the following structural constraints:
\begin{equation}
    s.t. \left\{ \begin{array}{l}
        {r_{\min }} \leq r \leq {r_{\max }}\\
        {S_{0,\min }} \leq {S_0} \leq {S_{0,\max }}\\
        \Delta {\theta _{\min }} \leq \Delta \theta  \leq \Delta {\theta _{\max }}
        \end{array} \right.   
\end{equation}


\begin{algorithm}
    \caption{Optimal Knee Joint Design Parameter Selection via High-Explosive Jump Control}
    \begin{algorithmic}[1]
        \Require A set of knee design parameters $\{P_1, P_2, \dots, P_n\}$
        \Ensure Maximum take-off energy and the corresponding structure parameter
        
        \State $max\_energy \gets -\infty$
        \State $optimal\_param \gets$ None
        
        \For{$i \gets 1$ to $n$}
            \State $energy \gets \text{simulateJump}(P_i)$ \Comment{Simulate high burst jump for structure parameter $P_i$ and record take-off energy}
            \If{$energy > max\_energy$}
                \State $max\_energy \gets energy$
                \State $optimal\_param \gets P_i$
            \EndIf
        \EndFor
        
        \State \Return $max\_energy, optimal\_param$
        \end{algorithmic}
\end{algorithm}

\subsection{Optimization Results and Analysis}
To guide the knee joint design for a full-scale humanoid robot, we set $l_1 = l_2 = 0.45$~m, $m_1 = 2.5$~kg, $m_2 = 5$~kg, and $m_3 = 20$~kg, assuming all links are rigid bodies with uniform mass distribution and centroids coinciding with their centers of mass. To ensure mechanical feasibility, the design parameters were set within the following ranges: $r$ from 25~mm to 75~mm, $S_0$ from 100~mm to 250~mm, and $\Delta\theta$ from $-3^\circ$ to $3^\circ$.  The knee actuator is a 72~V TQ-8526sp motor, with output characteristics shown in Fig.~\ref{fig:high_power_loss_region}.

\begin{table}[htbp]
  \centering
  \caption{Optimized Joint Parameters for Explosive Jumping Performance}
  \label{tab:optimized_params}
  \begin{threeparttable}
    \begin{tabularx}{\columnwidth}{
      >{\centering\arraybackslash}c
      >{\centering\arraybackslash}c
      >{\centering\arraybackslash}X
      >{\centering\arraybackslash}c
    }
      \toprule
      \makecell[c]{Joint Type}
        & \makecell[c]{Initial Angle\\(rad)}
        & \makecell[c]{Optimal Parameters \\\(\mathrm{(r,S_0,\Delta\theta)\;/\;k}\)}
        & \makecell[c]{Jump Height\tnote{a}\\\(H\) (m)} \\
      \midrule
      \multirow[c]{3}{*}{EVRR-K}
        & \(-2.6180\) & \((47,150,0)\) & 0.62 \\
        & \(-2.2689\) & \((47,150,0)\) & 0.51 \\
        & \(-1.9199\) & \((49,150,0)\) & 0.37 \\
      \midrule
      \multirow[c]{3}{*}{FRR-K}
        & \(-2.6180\) & 22             & 0.47 \\
        & \(-2.2689\) & 22             & 0.40 \\
        & \(-1.9199\) & 23             & 0.34 \\
      \bottomrule
    \end{tabularx}

    \begin{tablenotes}
      \footnotesize
      \item[a] Jump height \(H\) obtained from numerical simulation.
    \end{tablenotes}
  \end{threeparttable}
\end{table}

\begin{figure}[t]  
       \centering
    \begin{subfigure}{0.32\columnwidth}
        \centering
        \includegraphics[width=\linewidth]{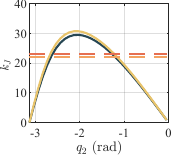}
        \caption{}
        \label{fig:best_ratio}
    \end{subfigure}
    \hfill
    \begin{subfigure}{0.32\columnwidth}
        \centering
        \includegraphics[width=\linewidth]{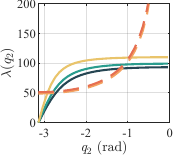}
        \caption{}
        \label{fig:best_general_ratio}
    \end{subfigure}
    \hfill
    \begin{subfigure}{0.32\columnwidth}
        \centering
        \includegraphics[width=\linewidth]{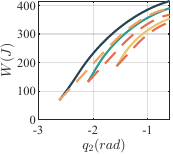}
        \caption{}
        \label{fig:best_Wm}
    \end{subfigure}

    \vspace{0.15cm}  

    \begin{subfigure}{0.32\columnwidth}
        \centering
        \includegraphics[width=\linewidth]{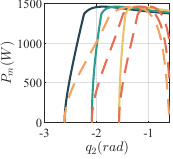}
        \caption{}
        \label{fig:best_Pm}
    \end{subfigure}
    \hfill
    \begin{subfigure}{0.32\columnwidth}
        \centering
        \includegraphics[width=\linewidth]{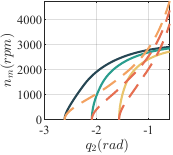}
        \caption{}
        \label{fig:best_nm}
    \end{subfigure}
    \hfill
        \begin{subfigure}{0.32\columnwidth}
        \centering
        \includegraphics[width=\linewidth]{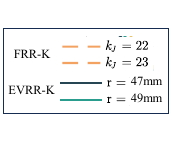}
        \label{fig:best_ratio_legend}
    \end{subfigure}

    \caption{Numerical simulation results for explosive jumping take-off with optimized EVRR-K and FRR-K. (a) Reduction ratio versus knee joint angle; (b) Transmission ratio from the knee motor to the CoM versus knee joint angle; (c) Work done by the motor versus; (d) Motor output power versus  knee joint angle; (e) Motor speed versus  knee joint angle.}
    \label{fig:best_results}
\end{figure}

To compare the jumping performance of EVRR-K and FRR-K under different initial knee joint angles at takeoff ($q_{2,0}$), we optimized the structural parameters for both types. The optimization results are summarized in Table~\ref{tab:optimized_params}. The theoretical takeoff height $H$ was calculated as
\begin{equation}\label{eq:takeoff_height}
H = \frac{W_{\text{takeoff}}}{m_{tot}g} - y_{\text{CoM,s}}
\end{equation}
where $y_{\text{CoM,s}}$ is the center of mass height with the leg fully extended. Results show that, across all initial angles, the EVRR-K consistently achieves significantly greater jumping heights than the FRR-K, with the most pronounced improvement reaching approximately 28.9\%.

The relationship between reduction ratio and joint angle for the optimal parameters is shown in Fig. \ref{fig:best_ratio}, where the reduction ratio decreases monotonically throughout the working range. The transmission ratio from the knee actuator to the center of mass (Fig. \ref{fig:best_general_ratio}) remains nearly constant at about 100 for the VRR joint, while for the FR joint it increases from 50 to over 200, indicating that the VRR design effectively mitigates the high-speed demand on the motor during the latter phase of takeoff.

Additionally, numerical simulation results (Figs. \ref{fig:best_Wm},\ref{fig:best_nm} and \ref{fig:best_results}) show that the EVRR-K performs more work during takeoff, with motor power rising rapidly at the start and decreasing gradually thereafter, while the FRR-K exhibits a slower power rise and a more rapid decline. The maximum motor speed of the EVRR-K joint motor remains below 3000~rpm in the latter stage, whereas the FRR-K motor requires 4000~rpm or higher, further demonstrating that the EVRR-K design effectively avoids operation in high power-loss regions.

\section{EXPERIMENTS}\label{sec5}
This section presents experimental validation of the proposed EVRR-K design paradigm. We conducted experiments on a one-degree-of-freedom (1-DOF) legged platform and on the full-scale humanoid robot BHR8-J1 to assess how the paradigm enhances knee motor performance during jumping and improves the robot’s jumping capabilities.

\subsection{1-DOF Leg-like Platform Experiments}
\subsubsection{Platform Design and Control}
To validate the method’s effectiveness and avoid interference from other joints, we constructed the 1-DOF platform shown in Fig.~\ref{fig:jump_small_platform_test} for vertical jump experiments. This platform was driven by a EVRR-K joint, with all other joints passive; its mathematical model is described in Section~\ref{sec2.2}. The main system parameters were $l_1 = l_2 = 0.42\,$m, $m_1 = 3.16\,$kg, $m_2 = 1.77\,$kg, and $m_3 = 20\,$kg. The EVRR-K parameters were obtained using the optimization method described in Section~\ref{sec4}, taking mechanical constraints into account; the final results were $S_0 = 0.259\,$m and $r = 0.047\,$m. The relationship between the reduction ratio and the knee angle is shown in Fig.~\ref{fig:best_ratio}. 
The actuator consists of a TQ-ILM8526sp motor, a 10 mm lead ball screw and an 18-bit encoder as shown in Fig.~\ref{fig:high_explosive_joint}. 

The control system included an NUC8 computer and an Elmo driver for real-time control, with a communication frequency of 1\,kHz. The jump control comprised three phases: takeoff, mid-air, and landing. During takeoff, a force-controlled PD algorithm adjusted the platform to the initial jump angle, after which the peak current ($I_q = 92\,$A) was applied until the takeoff-detection angle was reached. During the mid-air and landing phases, PD control regulated the output torque to maintain the joint at the set angle.

\subsubsection{Experimental Results and Analysis}
\begin{figure}[!h]
    \centering
    \includegraphics{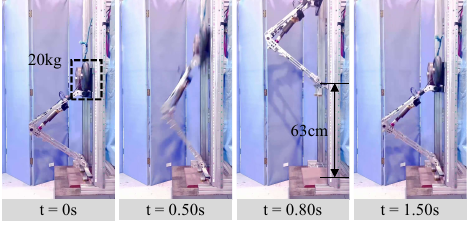}
    \caption{Vertical jump experiment on the one-DOF leg-like platform, actuated solely by the high-explosive primary knee joint (with all other joints passive). The knee joint is driven by a 72~V TQ-8526sp motor, with a peak power of 1.5~kW and a peak torque of 9.37~Nm.}
    \label{fig:jump_small_platform_test}
\end{figure}

\begin{figure}[t]
    \centering
    \begin{subfigure}[t]{0.4\linewidth}
        \centering
        \includegraphics{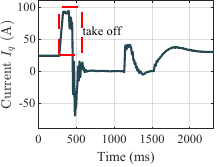}
        \caption{}
        \label{fig:small_platform_Iq}
    \end{subfigure}
    \hspace{0.05\linewidth}
    \begin{subfigure}[t]{0.4\linewidth}
        \centering
        \includegraphics{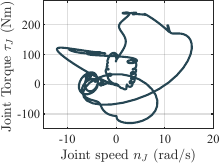}
        \caption{}
        \label{fig:small_platform_ratio}
    \end{subfigure}

    \vspace{0.4cm}

    \begin{subfigure}[t]{0.4\linewidth}
        \centering
        \includegraphics{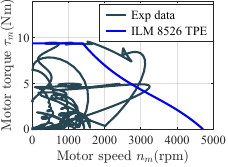}
        \caption{}
        \label{fig:small_platform_joint_speed}
    \end{subfigure}
    \hspace{0.05\linewidth}
    \begin{subfigure}[t]{0.4\linewidth}
        \centering
        \includegraphics{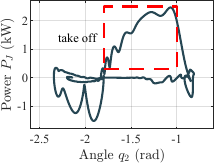}
        \caption{}
        \label{fig:small_platform_joint_power}
    \end{subfigure}

    \caption{Experimental data of explosive vertical jumping motion on a one-degree-of-freedom (1-DOF) leg-like platform. (a) Motor command  current; (b) Joint torque versus speed; (c) motor torque versus speed; (d)Knee angle versus joint power.}
    \label{fig:small_jumpExp_fig}
\end{figure}

The platform, with a total weight of 24.93\,kg (including a 20\,kg load), is approximately half the weight of a full-size robot. Under these conditions, the single knee joint achieved a jump height of 63\,cm (Fig.~\ref{fig:jump_small_platform_test}), demonstrating the effectiveness of the proposed method in enhancing jumping performance. To analyze the joint motor’s output characteristics, the $q$-axis current $I_q$ was recorded by the Elmo driver, and the motor position and speed were obtained from the encoder. The joint torque, angular velocity, and output power were then calculated as $\tau_{J} = \tau_{m} k(q_2)$ (where $\tau_{m} = K_{T} I_{q}$), $\omega_{J} = \omega_{m}/k(q_2)$, and $P_{J} = P_{m,\text{out}} = \tau_{J} \omega_{J}$, respectively. The results are shown in Fig.~\ref{fig:small_jumpExp_fig}.

Fig.~9 presents the data from a complete jump experiment. Fig.~9(a) shows both the motor $q$-axis current and the command current. The actual current closely follows the command throughout most of the motion, indicating effective current tracking; however, in the take-off stage (highlighted by the red box), current tracking degrades in the later phase as the increasing motor speed reduces the torque output capability. This demonstrates the effectiveness of the proposed control method. As shown in Fig.~9(b), the peak torque reached 250\,Nm, and the peak speed reached 15\,rad/s, highlighting the high torque and speed demands of the jumping task. According to the motor torque--speed trajectory in Fig.~9(c), the motor operated predominantly within its performance envelope, with the maximum speed around 3000\,rpm, thereby avoiding the high power loss region. Fig.~9(d) shows that the joint power rapidly increased to the peak output of 1.5\,kW during take-off and continued to rise until approaching the take-off angle, reflecting the capability of the proposed method to quickly achieve and sustain high-power output.

In addition, as shown in Fig.~9(d), during the late stage of take-off, the motor output power exceeded its rated peak value, and the motor operation surpassed the manufacturer’s specified performance envelope (Fig.~9(c)). This can be attributed to the motor’s inherent overload capability, although control stability deteriorates beyond this limit. Furthermore, fluctuations in the frictional force along the track led to an irregular power increase in the early phase of take-off.

\subsection{Full-Scale Humanoid Robot Jumping Experiment}

\begin{figure*}[!t]
    \centering
    \begin{subfigure}{0.32\textwidth}
        \centering
        \includegraphics{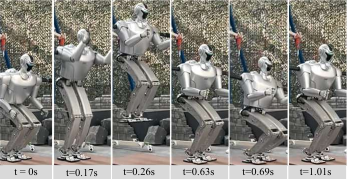}
        \caption{Height jump (0.5m).}
        \label{fig:full-scale-jumpExp_subfig1}
    \end{subfigure}
    \hfill
    \begin{subfigure}{0.32\textwidth}
        \centering
        \includegraphics{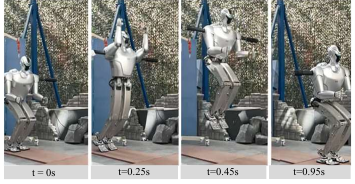}
        \caption{forward  jump (1.1m).}
        \label{fig:full-scale-jumpExp_subfig2}
    \end{subfigure}
    \hfill
    \begin{subfigure}{0.32\textwidth}
        \centering
        \includegraphics{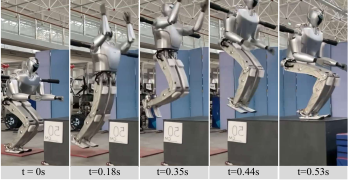}
        \caption{Box jump (0.5m ).}
        \label{fig:full-scale-jumpExp_subfig3}
    \end{subfigure}

    \caption{Full-scale humanoid robot jump experiments.}
    \label{fig:full-scale-jumpExp}
\end{figure*}
\begin{figure}[!h]
\centering
\includegraphics[width=0.7\linewidth]{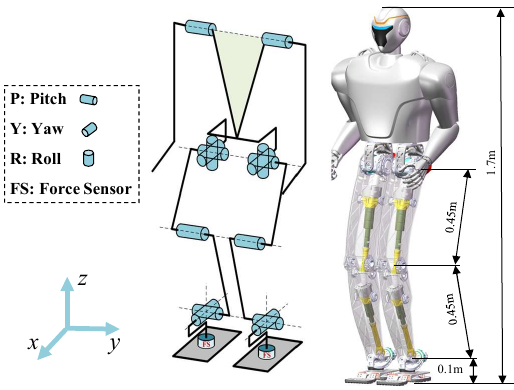}
\caption{The full-scale BHR8-J1 humanoid robot equipped with the proposed EVRR-K and its degree-of-freedom configuration.}
\label{fig:exp_jump_robot}
\end{figure}
\begin{figure}[!h]
    \centering
       \begin{subfigure}{0.23\textwidth}
        \centering
        \includegraphics[width=\textwidth]{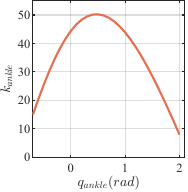}
        \caption{}\label{fig:joint_diagram}
    \end{subfigure}
    \hfill
    \begin{subfigure}{0.23\textwidth}
        \centering
        \includegraphics[width=\textwidth]{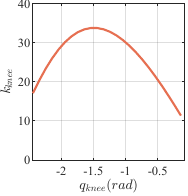}
        \caption{}\label{fig:lactuator_structure}
    \end{subfigure}
    \caption{Reduction ratios of the knee and ankle joints in BHR8-J1. (a) Ankle joint reduction ratio; (b) Knee joint reduction ratio.}
    \label{fig:reduction_ratios}
 \end{figure}

To further validate the effectiveness of EVRR-K in a robotic context, we conducted jump tests on the full-scale humanoid robot BHR8-J1 equipped with the proposed EVRR-K (Fig.~\ref{fig:exp_jump_robot}). The robot has 14 degrees of freedom, with three at each hip, one at each knee, and two at each ankle, as detailed in Table~\ref{tab:actuator_config}. The reduction ratios for the knee and ankle joints are shown in Fig.~\ref{fig:reduction_ratios}. The total mass is 45\,kg. The control system comprises an NUC8 computer and 14 Elmo drivers for real-time control over EtherCAT with a 1\,ms cycle. Each motor is fitted with an 18-bit encoder, and the supply voltage is 72\,V.

\begingroup
\setlength{\tabcolsep}{4pt} 
\small
\begin{table}[t]
\centering
\caption{Actuation configurations for BHR8-J1 joints}
\label{tab:actuator_config}
\begin{tabular}{@{}l l l l@{}}
\toprule
Joint (Axis)      & Actuator Type           & Ratio / Lead     & Motor Model    \\
\midrule
Hip (Yaw)         & Harmonic drive          & 100:1            & TQ-ILM5014sp   \\
Hip (Roll)        & Harmonic drive          & 100:1            & TQ-ILM8513sp   \\
Hip (Pitch)       & Planetary gearbox       & 24.695:1         & TQ-ILM8523sp   \\
Knee (Pitch)      & Ball screw              & 10 mm lead       & TQ-ILM8526sp   \\
Ankle (Pitch)     & Ball screw              & 5 mm lead        & TQ-ILM7018sp   \\
Ankle (Roll)      & Harmonic drive          & 50:1             & TQ-ILM5014sp   \\
\bottomrule
\end{tabular}
\end{table}
\endgroup

BHR8-J1 successfully performed a 0.5\,m vertical jump, a 1.1\,m forward jump, and a 0.5\,m box jump, demonstrating strong jumping capabilities, as shown in Fig.~\ref{fig:full-scale-jumpExp}. In addition, the box jump results indicate that neither the vertical nor the forward jump reached their optimal performance, suggesting considerable potential for further improvement. The detailed jumping control and trajectory optimization methods are described in~\cite{qi2023vertical}. 

\begin{figure}[htbp]
    \centering
    \begin{subfigure}[t]{\columnwidth}
        \centering
        \includegraphics[width=0.95\columnwidth]{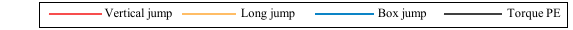}
    \end{subfigure}
    \vspace{0.1cm}

    \begin{subfigure}{0.95\columnwidth}
        \centering
        \includegraphics[width=\linewidth]{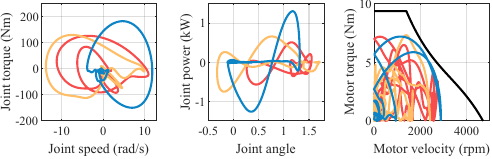}
        \caption{Data of hip joint.}
        \label{fig:jumpExp_hip}
    \end{subfigure}
    \vspace{0.15cm}

    \begin{subfigure}{0.95\columnwidth}
        \centering
        \includegraphics[width=\linewidth]{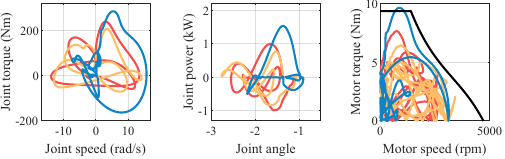}
        \caption{Data of knee joint.}
        \label{fig:jumpExp_knee}
    \end{subfigure}
    \vspace{0.15cm}

    \begin{subfigure}{0.95\columnwidth}
        \centering
        \includegraphics[width=\linewidth]{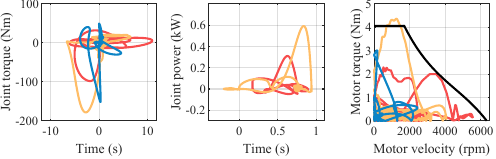}
        \caption{Data of ankle joint.}
        \label{fig:jumpExp_ankle}
    \end{subfigure}

    \caption{Hip, knee, and ankle pitch direction data for the 50cm high jump, 1.1m forward jump, and 50cm box jump.}
    \label{fig:overall}
\end{figure}

To further validate the effectiveness of the proposed method, motor current data were recorded from the Elmo drivers, along with position and velocity data from the encoders. The torque, speed, and power characteristics of the hip, knee, and ankle joints during various jumping tasks were subsequently calculated, as shown in Fig.~\ref{fig:overall}.

As shown in Fig.~\ref{fig:overall}, among the hip, knee, and ankle joints, the knee consistently demands the highest torque, speed, and power across all jumping tasks. For the hip and ankle (Figs.~\ref{fig:jumpExp_hip} and \ref{fig:jumpExp_ankle}), peak torque stays below 200\,Nm, peak speed below 12\,rad/s, and peak power under 1.2\,kW. In contrast, during the 0.5\,m box jump (Fig.~\ref{fig:jumpExp_knee}), the knee reaches 286\,Nm, 15.5\,rad/s, and 1.5\,kW (motor peak at 72\,V). Thanks to the variable reduction ratio, the knee motor speed remains under 3000\,rpm, avoiding the high power-loss region. A fixed-ratio joint would require a ratio of at least 30 to match the same torque, forcing motor speeds to exceed 4500\,rpm, which is near no-load speed and enters the high power-loss region. The proposed design thus lowers the required motor speed, reduces power loss, and improves efficiency.

Furthermore, the robot did not demonstrate sustained high-power output during the jumping process, mainly because it was not operating at its maximum capacity. In addition, the current control and trajectory optimization method~\cite{qi2023vertical} does not fully exploit the advantages of the variable reduction ratio. Future work will focus on leveraging the benefits of the variable reduction ratio to further enhance the robot's jumping performance.

\section{Conclusion and Future Work}\label{sec6}
In this paper, we proposed a novel knee joint design paradigm employing a dynamically decreasing reduction ratio to enhance the explosive power output and jumping performance of humanoid robots. By coupling the reduction ratio to the joint angle, our approach enables sustained high-power output throughout the jump phase, effectively overcoming the limitations imposed by conventional fixed-ratio or minimally varying transmission designs. The compact and efficient linear actuator-driven guide-rod mechanism was developed to realize the proposed strategy, and an optimization framework guided by explosive jump control was introduced to maximize performance. Experimental validation demonstrated that the proposed approach achieves a 63cm vertical jump on a single-joint test platform, representing a 20\% improvement over the optimal fixed-ratio design. Furthermore, integration into a full-scale humanoid robot enabled jumps of 1.1m in length and 0.5~m in height, confirming the effectiveness and practical applicability of the proposed methodology.

However, the crucial role of the ankle joint in humanoid jumping dynamics remains unaddressed in this study. Future work will focus on developing and integrating advanced ankle joint technologies to further improve the explosive jumping performance and agility of humanoid robots.


\bibliographystyle{IEEEtran}

\bibliography{references}













\end{document}